\documentclass[preprint]{elsarticle}

\usepackage[utf8]{inputenc}
\usepackage{xcolor}
\usepackage[title]{appendix}
\usepackage[letterpaper,top=2cm,bottom=2cm,left=2cm,right=2cm,marginparwidth=1.75cm]{geometry}

\usepackage{amsmath}
\usepackage{palatino,graphicx,wrapfig}
\usepackage[small,it]{caption}
\usepackage{amsmath}
\usepackage{amssymb,bm}
\usepackage{parskip}
\usepackage{multirow}
\usepackage{booktabs}

\usepackage[
linesnumbered,ruled,vlined]{algorithm2e}
\usepackage {algpseudocode}
\usepackage{algorithmicx}
\usepackage{algcompatible}

\journal{}

\title{Interpretable Diagnostics and Adaptive Data Assimilation for Neural ODEs via Discrete Empirical Interpolation}

\author[label1]{Hojin Kim}
\author[label1,label2]{Romit Maulik}

\affiliation[label1]{organization={School of Mechanical Engineering, Purdue University},
            city={West Lafayette},
            postcode={47907},
            state={IN},
            country={USA}}

\affiliation[label2]{organization={Mathematics and Computer Science Division, Argonne National Laboratory},
            city={Lemont},
            postcode={60439},
            state={IL},
            country={USA}}

\date{Jan 2026}

\begin{document}

\begin{abstract}

We present a framework that leverages the Discrete Empirical Interpolation Method (DEIM) for interpretable deep learning and dynamical system analysis. Although DEIM efficiently approximates nonlinear terms in projection-based reduced-order models (POD-ROM), its fixed interpolation points are repurposed for identifying dynamically representative spatial structures in learned models. We apply DEIM as an interpretability tool to examine the learned dynamics of a pre-trained Neural Ordinary Differential Equation (NODE) for two-dimensional vortex-merging and backward-facing step flows. DEIM trajectories reveal physically meaningful structures in NODE predictions and expose failure modes when extrapolating to unseen flow configurations. Building on this diagnostic capability, we further introduce a DEIM-guided data assimilation strategy that injects sparse, dynamically representative corrections into the NODE rollout. By allocating a limited nudging budget to DEIM-identified sampling locations, the framework significantly improves long-term stability and predictive accuracy in out-of-distribution scenarios for the two-dimensional vortex-merging flow. Additional experiments for a flow over a backward-facing step reveal regime-dependent gains, with alternative sampling strategies performing competitively as well. These results demonstrate that DEIM can serve as an interpretable diagnostic and control framework for understanding and enhancing neural differential equation models.
\end{abstract}


\maketitle

\section{Introduction}

Machine learning has recently emerged as a powerful paradigm for modeling high-dimensional nonlinear dynamical systems. In particular, neural differential equation models such as Neural Ordinary Differential Equations (NODEs) provide a continuous-time formulation for learning the evolution of physical systems directly from data \cite{chen2018neural}. More broadly, scientific machine learning has developed a rich class of surrogate models for PDE-governed dynamics, including operator-learning architectures such as DeepONet, neural operators, and Fourier neural operators, as well as geometry-aware and graph-based simulators for unstructured meshes \cite{lu2021learning, kovachki2023neural, li2020fourier, li2023fourier, pfaff2021learning}. These approaches have substantially expanded the range of flows and transport phenomena that can be approximated from data, often with impressive short-term predictive accuracy. However, they still often operate as black boxes, offering limited interpretability and uncertain generalization when applied to unseen physical regimes. In addition, long-horizon rollouts can be highly sensitive to small modeling errors; recent studies on stabilized NODEs and refinement-based neural PDE surrogates continue to highlight the difficulty of maintaining phase accuracy, stability, and coherent structure over extended prediction horizons \cite{linot2023stabilized, lippe2023pde}. Developing interpretable frameworks that can diagnose, analyze, and improve learned dynamical models is therefore essential for trustworthy scientific machine learning.

A common line of work toward trustworthy surrogate modeling is to incorporate physical structure---such as invariances, conservation laws, or stability-promoting inductive biases---and to design post hoc diagnostics that expose learned failure modes. In parallel, recent work on interpretable scientific machine learning has emphasized that surrogate models should not only be accurate, but also diagnostically informative, so that one can identify which spatial regions or latent mechanisms are most responsible for prediction and out-of-distribution degradation \cite{arzani2025interpreting, barwey2025interpretable, raut2025fignn}. These developments suggest that interpretability is not merely an explanatory add-on; it can also serve as a mechanism for localizing model deficiencies and guiding corrective interventions.

Projection-based reduced-order modeling (ROM) methods, such as Proper Orthogonal Decomposition (POD), have long provided physically interpretable low-dimensional descriptions of complex flows \cite{rowley2017model}. Within this literature, the Discrete Empirical Interpolation Method (DEIM) identifies sparse interpolation points for efficiently approximating nonlinear terms, thereby linking reduced-order dynamics to spatially localized physical structures \cite{chaturantabut2010nonlinear, cstefuanescu2015pod}. Subsequent variants, such as Q-DEIM, further refined the point-selection procedure, while related sparse-sensing and sensor-placement studies showed that low-rank libraries and DEIM-type nonlinear libraries can be exploited for state reconstruction, regime identification, and optimized measurement allocation \cite{drmac2016new, manohar2018data, sargsyan2015nonlinear}. Although DEIM is traditionally introduced as a hyper-reduction tool for ROMs, these developments suggest a broader interpretation: DEIM can also be viewed as an interpretable sparse sampling operator that highlights dynamically representative spatial locations.

This perspective naturally connects DEIM to data assimilation (DA), where sparse observations are used to stabilize and correct imperfect models. Classical DA approaches range from nudging and relaxation methods to ensemble- and Kalman-filter-based updates \cite{azouani2014continuous, evensen2003ensemble}. More recently, learning-based filtering and prediction-correction frameworks, such as KalmanNet and neural-operator-based DA, have shown that data-driven surrogates can also benefit substantially from online observational feedback \cite{revach2022kalmannet, singh2024learning, cheng2024efficient}. In practical fluid settings, however, the number of available observations is often severely limited, and the effectiveness of DA depends critically on \emph{where} those observations are placed. Under a fixed observation budget, spatial allocation can be as important as the assimilation rule itself, particularly in regimes with coherent vortical structures, moving interfaces, or localized separation zones \cite{mons2017optimal, deng2021deep, xu2024optimal, sashittal2021data, mons2021linear}.

In this work, we revisit DEIM from this broader diagnostic and control-oriented perspective, and apply it to the analysis of learned NODE dynamics in two-dimensional vortex-merging and backward-facing step flows. Our starting hypothesis is that dynamically identified DEIM points correspond to physically important locations in the flow field and can therefore be used to probe the internal behavior of a learned surrogate. To test this idea, we apply DEIM to \emph{time-windowed} snapshots of the NODE-predicted right-hand side and extract trajectories of representative sampling points. We then compare these trajectories with those obtained from ground-truth dynamics, using their agreement or divergence as an interpretable signature of the NODE's capacity to generalize to unseen flow conditions.

Leveraging this diagnostic signal, we further introduce a DEIM-guided nudging-based DA strategy. Rather than distributing sparse corrections uniformly across the domain, we allocate the nudging budget to dynamically representative locations identified by DEIM and enrich these locations through kernel density estimation (KDE). This yields a structure-aware observation policy that improves long-term stability and predictive accuracy while remaining physically interpretable. Our experiments also reveal regime-dependent behavior: sampling guided by coherent DEIM trajectories is particularly effective for vortex interaction dynamics, whereas advection-dominated separated flows may favor more instantaneous or spatially distributed observation policies.

Taken together, these components position DEIM not only as a model reduction method, but also as a bridge between interpretable diagnostics, sparse sensing, and observation placement for the online correction of learned dynamical systems.

The main contributions of this work are as follows:
\begin{itemize}
\item We propose a time-windowed DEIM procedure to extract trajectories of representative sampling points from learned NODE dynamics and compare them with trajectories derived from ground-truth dynamics.
\item We show that DEIM trajectories provide an interpretable diagnostic signal that correlates with NODE rollout failure modes in extrapolative flow configurations, including vortex-merging and backward-facing step flows.
\item We introduce a DEIM-guided nudging-based DA strategy, in which a limited nudging budget is allocated to DEIM-selected and KDE-enriched locations, improving stability and accuracy under sparse observations.
\item We perform controlled comparisons against alternative sparse nudging strategies (random, uniform, and Top-RHS selection) and analyze regime-dependent behavior across vortex-merging and backward-facing step flows.
\end{itemize}

\section{Background}

\subsection{Projection-based Reduced-order Models}
Proper Orthogonal Decomposition (POD) is a data-driven method to construct low-dimensional subspaces that optimally capture the dynamics of high-dimensional systems. This process includes two steps: snapshot collection and basis construction. Snapshot collection extracts solutions $\mathbf{u}(t_i) \in \mathbb{R}^{n}$ from full-order simulations or experiments at time instances $t_i(i=1,...,N)$. Then, a basis is identified via the singular value decomposition (SVD) on the snapshot matrix $\mathbf{U} = [\mathbf{u}(t_1), ..., \mathbf{u}(t_N)] \in \mathbb{R}^{n\times N}$ to obtain orthogonal POD modes $\mathbf{\Psi} = [\Psi_1, ..., \Psi_m] \in \mathbb{R}^{n\times m}$, where $m$ is the number of truncated POD modes and $n>>m$. $\Psi_i (i=1,...,m)$ is ranked by each mode's energy content. The number of truncated POD modes is typically determined by setting an energy threshold, such that the retained modes collectively capture a predetermined percentage (e.g., 99\%) of the total variance (energy) in the dataset. This is achieved by computing the cumulative sum of the eigenvalues (or singular values) and selecting the smallest number of modes for which the cumulative energy exceeds the chosen threshold. In some cases, additional dynamical or physical considerations may influence the final selection of modes. 

Using the orthonormal property of these POD modes, Galerkin projection projects governing equations (e.g. nonlinear PDEs) onto a reduced subspace spanned by these modes. A nonlinear PDE in Eq. \ref{eqn:nonlinear_PDE},

\begin{equation}
\label{eqn:nonlinear_PDE}
\frac{\partial \mathbf{u}}{\partial t} = \mathcal{L}_f[\mathbf{u}] + \mathcal{N}_f[\mathbf{u}]
\end{equation}

, where $\mathcal{L}_f$ and $\mathcal{N}_f$ are linear and nonlinear operators in the full-order space, respectively, can be approximated using the POD modes as follows:

\begin{equation}
\label{eqn:Galerkin_1}
\frac{\partial \mathbf{\Psi}\mathbf{a}}{\partial t} = \mathcal{L}_f[\mathbf{\Psi}\mathbf{a}] + \mathcal{N}_f[\mathbf{\Psi}\mathbf{a}]
\end{equation}
, where $\mathbf{a} = [a_1,...,a_m] \in \mathbb{R}^{m} $ is temporal POD coefficients, and $\mathbf{u}$ is approximated as $\mathbf{u}\approx\mathbf{\Psi}\mathbf{a}=\sum_{i=1}^{m}{\psi_ia_i}$

Then, by using the orthonormality of the POD modes, the final reduced system becomes as follows:

\begin{equation}
\label{eqn:Galerkin_2}
\frac{\partial \mathbf{a}}{\partial t} = \mathcal{L}_r[\mathbf{a}] + \mathbf{\Psi}^T\mathcal{N}_f[\mathbf{\Psi}\mathbf{a}]
\end{equation}

, where $\mathcal{L}_r$ is a linear operator in the reduced-order space. This approach preserves system dynamics while reducing computational cost, because the dimension of $\mathbf{a}$ in the reduced system, $m$, is orders of magnitude smaller than that of $\mathbf{u}$, $n$ in the original full-order system. However, since the POD modes do not commute with the nonlinear operator, unlike the linear operator, $\mathcal{N}_f$ remains expensive to evaluate in high dimensional systems.

\subsection{Discrete empirical interpolation method (DEIM)}

In order to address the complexity of calculating the nonlinear term in Eq. \ref{eqn:Galerkin_2}, Discrete Empirical Interpolation Method (DEIM) \cite{chaturantabut2010nonlinear} approximates $\mathcal{N}_f[\mathbf{u}]$ using a sparse subset of interpolation points. This method starts by constructing a separate POD basis $\Phi \in \mathbb{R}^{n \times l}$ from nonlinear snapshots, $\mathbf{U}_{nl} = [\mathbf{u}_{nl}(t_1), ..., \mathbf{u}_{nl}(t_N)] \in \mathbb{R}^{n\times N}$. Here, $l$ is the number of truncated POD modes for nonlinear snapshots, and $n>>l$. Then, the DEIM sampling matrix, $\mathbf{P}=[e^{p_1}, ...,e^{p_l}]$, where $\mathbf{P} \in \mathbb{R}^{n \times l}$ and $e^{p_k}(k=1,...,l)$ is a one-hot vector with a 1 at the $p_k$-th entry and zeros elsewhere, is constructed using a greedy algorithm to select $l$ interpolation points that minimize the approximation error. Then, $\mathcal{N}_f[\mathbf{u}]$ can be approximated as follows:

\begin{equation}
\label{eqn:DEIM}
\mathcal{N}_f[\mathbf{u}] \approx \Phi(\mathbf{P}^T\Phi)^{-1}\mathbf{P}^T\mathcal{N}_f[\mathbf{u}]
\end{equation}

Eq. \ref{eqn:DEIM} reduces the cost of evaluating $\mathcal{N}_f[\mathbf{u}]$ from $O(n)$ to $O(l)$, because the nonlinear terms $\mathcal{N}_f[\mathbf{u}]$ are only calculated at $x_{p_k} (k=1,...,l)$. The sampling matrix $\mathbf{P}$ is pre-computed and fixed during the online stage of the ROM. Eq. \ref{eqn:DEIM} is substituted into Eq. \ref{eqn:Galerkin_2} to get the final form of the PDE in the reduced order space as follows:

\begin{equation}
\label{eqn:Galerkin-DEIM}
\frac{\partial \mathbf{a}}{\partial t} = \mathcal{L}_r[\mathbf{a}] + \mathbf{\Psi}^T\Phi(\mathbf{P}^T\Phi)^{-1}\mathbf{P}^T\mathcal{N}_f[\mathbf{u}]
\end{equation}

DEIM can reduce the computational cost of POD-Galerkin frameworks by maintaining a low computational complexity for the calculation of the nonlinear term. For details on the DEIM algorithm, the reader is referred to the previous work \cite{chaturantabut2010nonlinear}.

% \subsection{Differentiable Physics}

% Differentiable physics frameworks aim to seamlessly integrate physical simulation and machine learning by enabling gradient-based optimization through the governing equations of dynamical systems \cite{sanderse2024scientific}. These frameworks reformulate numerical solvers in a differentiable manner, allowing model parameters to be optimized via backpropagation while preserving the physical consistency imposed by the underlying PDEs. Differentiability can be achieved using automatic differentiation, adjoint-based methods, or source-to-source transformations of existing solvers. Recent studies have leveraged such frameworks to train machine learning models for LES turbulence closures, reduced-order model corrections, and boundary conditions in fluid–structure interaction problems \cite{sirignano2023deep, kim2023generalizable, ahmed2023multifidelity, fan2024differentiable}. Ultimately, differentiable physics bridges the gap between data-driven and physics-based modeling, providing a unified foundation for interpretable and physically consistent learning.

\subsection{Neural Ordinary Differential Equations}
A neural ordinary differential equation (NODE) \cite{chen2018neural} can be formulated as:
\begin{equation}
\label{eqn:NODE}
    \frac{d\mathbf{u(t)}}{dt} = f(\mathbf{u(t)}, t ;\theta)
\end{equation}
, where $\mathbf{u(t)} \in \mathbb{R}^N$ is the state at each time step, and $f(\mathbf{u(t)}, t, \theta)$ represents the hidden dynamics parameterized by a neural network with parameters $\theta$. The solution is then obtained by integrating this dynamics with an ODE solver. For a nonlinear PDE as in Eq. \ref{eqn:nonlinear_PDE}, learning $f(\mathbf{u(t)}, t, \theta)$ using $\mathbf{u(t)}$ implies that one learns the entirety of the right-hand side of the PDE including the linear operator $\mathcal{L}_f$ and the nonlinear operator $\mathcal{N}_f$, which is evaluated at each time step. This can be expressed as:
\begin{equation}
\label{eqn:NODE-PDE}
    \frac{d\mathbf{u(t)}}{dt} = \mathcal{L}_f[\mathbf{u(t)}] + \mathcal{N}_f [\mathbf{u(t)}] \approx f(\mathbf{u(t)}, t ;\theta)
\end{equation}

\section{DEIM for Analysis of Trained Neural ODEs}
In this section, we perform an interpretability analysis of the pre-trained NODEs. Specifically, windowed-DEIM is applied to the learned dynamics of the PDEs, $f(\mathbf{u(t)}, t ;\theta)$ in Eq. \ref{eqn:NODE-PDE}, to identify representative sampling points and examine their dynamic behavior. 

\subsection{Problem Setup}

\subsubsection{Two-Dimensional Vortex-Merging}

The first test case is the two-dimensional vortex-merging problem \cite{ahmed2023physics}. The incompressible two-dimensional Navier–Stokes equations, expressed in terms of the vorticity and stream function, can be written as follows:

\begin{equation}
\label{eqn:vortex-merging}
\frac{\partial \omega}{\partial t} = -J(\omega, \psi) + \frac{1}{Re}\nabla^2\omega
\end{equation}
\begin{equation}
\nabla^2\psi=-\omega
\end{equation}
\begin{equation}
J(\omega, \psi)=\frac{\partial \omega}{\partial x} \frac{\partial \psi}{\partial y} - \frac{\partial \omega}{\partial y} \frac{\partial \psi}{\partial x },
\end{equation}
 
where $\omega$ and $\psi$ are vorticity and stream function, respectively, and $J(\omega, \psi)$ is the Jacobian operator, which determines the nonlinear advection of vorticity. The NODE is trained to learn the right-hand side of Eq. \ref{eqn:vortex-merging}, which includes both the nonlinear Jacobian operator and the linear viscous dissipation term.

\begin{figure}
  \begin{center}
      \includegraphics[width=\linewidth]{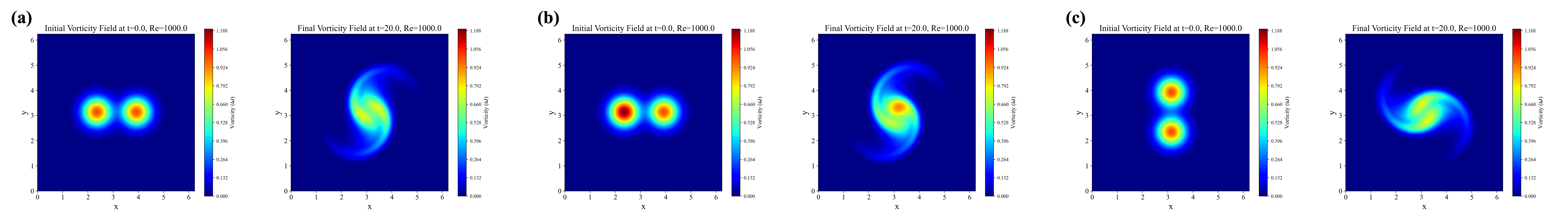}
  \end{center}
  \caption{Final solutions of the two-dimensional vortex-merging problem for 
$Re=1,000$ with different initial conditions: (a) horizontally symmetric vortices, (b) horizontally asymmetric vortices, and (c) vertically symmetric vortices}
  \label{fig:vortexmerging_inital_final}
\end{figure}

The computational domain is set to $[0, 2\pi]\times[0, 2\pi]$, discretized on a uniform Cartesian grid with $N_x=N_y=128$ points. The timestep size is $dt = 1\times10^{-2}$, with a total of 2000 timesteps, and the SSP-RK3 scheme is used for time integration. We use the first 1500 snapshots from the solution initialized with horizontal vortices, as shown in Fig.~\ref{fig:vortexmerging_inital_final}(a), for training, and apply the trained model to the remaining initial conditions. Fig. \ref{fig:vortexmerging_inital_final} presents the final solutions of Eq. \ref{eqn:vortex-merging} for different initial conditions at $Re=1,000$.

\paragraph{Neural ODE Architecture}
Because the vortex-merging problem is defined on a uniform Cartesian grid, we employ a convolutional neural network (CNN) as the backbone of the Neural ODE. The state variable is the scalar vorticity field $\omega(t) \in \mathbb{R}^{128 \times 128}$, and the CNN maps this field to the predicted temporal derivative $\dot{\omega}(t)$. The architecture consists of four convolutional layers with $3\times3$ kernels, 32 feature channels, and ReLU activations, each followed by a residual connection that propagates low-level spatial information through the network. A final $1\times1$ convolutional layer projects the 32 intermediate feature channels back to a single scalar field. This design allows the NODE to efficiently capture local nonlinear interactions and spatial correlations in the vortex dynamics.

\paragraph{Training Details}
The CNN-based NODE is trained by minimizing a multi-step rollout loss on trajectory segments of length $L=60$. Given a start time index $t_0$, the initial condition is set to $\omega(t_0)$, and the Neural ODE is integrated forward using the SSP-RK3 scheme over $L$ rollout steps to obtain predictions $\hat{\omega}(t_0+k)$ for $k=1, 2, \dots, L$. The training objective is the mean-squared error over the rollout segment,
\begin{equation}
\mathcal{L}(\theta)=\frac{1}{L}\sum_{k=1}^{L}
\left\|\hat{\omega}(t_0+k)-\omega(t_0+k)\right\|_2^2,
\end{equation}
where the norm is evaluated over all grid points. At each training iteration, a batch of five initial conditions is randomly sampled from the training interval. The dataset is split along the temporal axis, using the first 1500 snapshots for training and the remaining 500 for validation. Training was performed using the Adam optimizer \cite{kingma2014adam} with a learning rate of $10^{-3}$ on a single NVIDIA A100 GPU on Penn State ICDS Roar Collab for 550 epochs, requiring approximately 10 hours in total.

\subsubsection{Two-Dimensional Backward-Facing Step}

The second test case is the flow over a two-dimensional backward-facing step (BFS), a common benchmark problem in fluid dynamics that has been widely studied in both experimental \cite{armaly1983experimental} and computational \cite{wee2004self} settings. BFS flow is characterized by flow separation at the step and a subsequent reattachment region further downstream, which introduces a recirculation region near the step and transient vortex shedding. The governing equations are the incompressible Navier–Stokes equations as follows:

\begin{align}
\frac{\partial \mathbf{u}}{\partial t}+\nabla\cdot(\mathbf{u}\otimes\mathbf{u})&=\frac{1}{Re}\nabla^2\mathbf{u}-\frac{1}{\rho}\nabla p+\mathbf{f} \\
\nabla\cdot\mathbf{u}&=0.
\label{eq:ns}
\end{align}

The NODE is trained to learn the right-hand side of the momentum equation, capturing the combined effects of advection, viscous diffusion, and the pressure gradient.

\begin{figure}
  \begin{center}
      \includegraphics[width=0.8\linewidth]{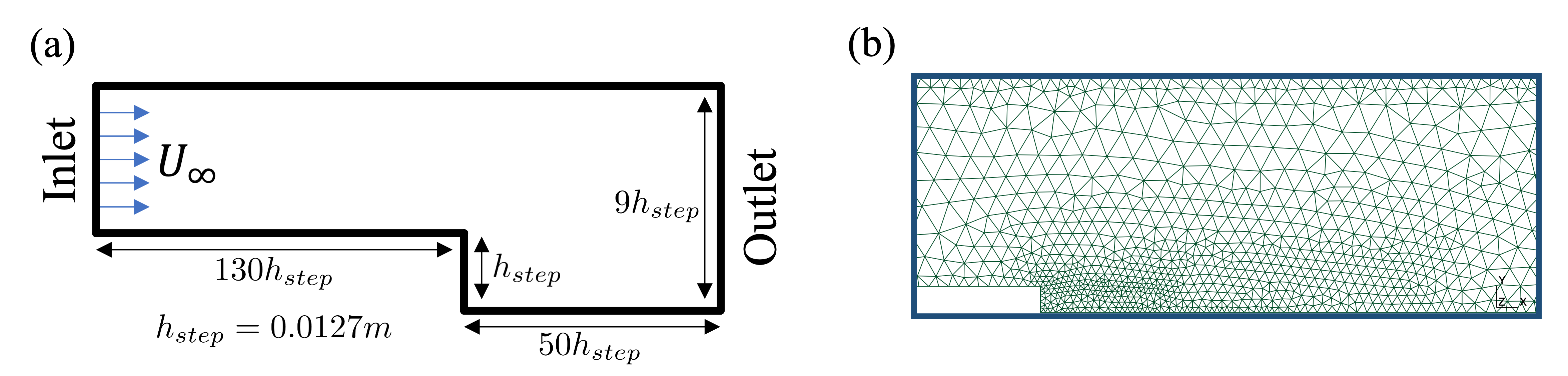}
  \end{center}
  \caption{(a) Schematic of the backward-facing step domain with inlet, outlet,
  and dimensions labeled. (b) A subsection of the computational mesh, where a
  higher density of elements is placed downstream of the step.}
  \label{fig:dom}
\end{figure}

For generating this dataset, we use the finite element method (FEM)-based solver FEniCS \cite{fenics} with the implicit pressure correction scheme (IPCS) \cite{goda1979multistep}. A schematic of the BFS configuration and its computational domain is presented in Fig.~\ref{fig:dom}. A total of 5,253 unstructured triangular cells are used for the computational mesh, with Taylor--Hood finite elements. As shown in Fig.~\ref{fig:dom}(b), a higher density of elements is placed downstream of the step in order to effectively capture flow separation and the trajectories of shed vortices. The timestep size is $dt = 1\times10^{-4}$, and the simulation data from $t=0.01$ to $t=0.05$ are used for model development. We split the time series along the temporal axis, using the first 80\% of the snapshots for training and the remaining 20\% for validation.

\begin{figure}
  \begin{center}
      \includegraphics[width=\linewidth]{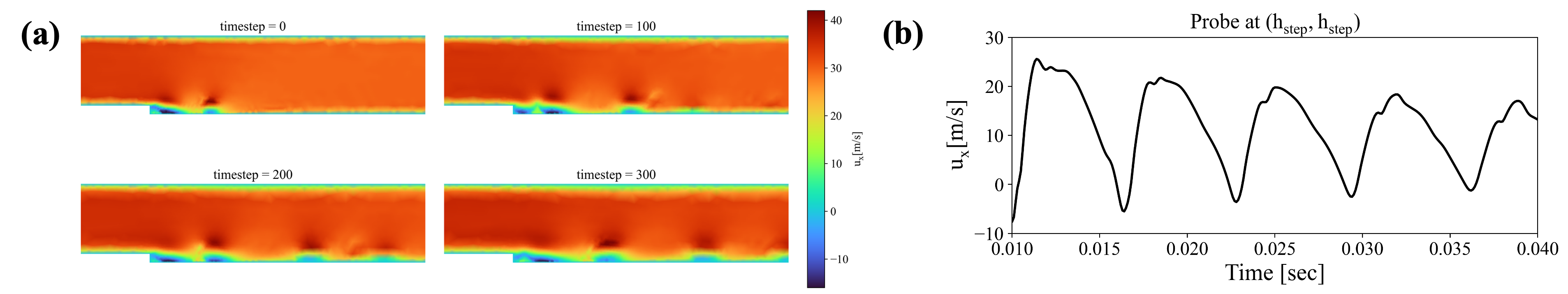}
  \end{center}
  \caption{(a) Streamwise velocity contours around the BFS at different timesteps. Here, the plots commence from the 100th snapshot after starting the simulation; (b) Pointwise streamwise velocity at the location $(x,y)=(h_{\mathrm{step}},h_{\mathrm{step}})$}
  \label{fig:BFS_contour}
\end{figure}

Figure \ref{fig:BFS_contour}(a) presents contours of the streamwise velocity around the BFS at different time instants. The flow exhibits periodic vortex formation induced by separation at the step edge, with the vortices subsequently convected downstream toward the channel outlet and gradually dissipated due to viscous effects. Figure \ref{fig:BFS_contour}(b) shows the temporal evolution of the pointwise streamwise velocity at $(x,y) = (h_{\mathrm{step}}, h_{\mathrm{step}})$, revealing a clear periodic fluctuation associated with the passage of shed vortices.

\paragraph{Neural ODE Architecture}
Because the BFS problem is defined on a highly nonuniform unstructured mesh, CNNs, which are naturally designed for uniform Cartesian grids, are not well suited. We therefore employ a graph neural network (GNN) as the backbone of the Neural ODE. GNNs have demonstrated their ability to model spatiotemporal data for fluid-flow prediction tasks on unstructured meshes \cite{pfaff2021learning, barwey2023multiscale, kim2024generalizabledatadriventurbulenceclosure, barwey2025interpretable}. The unstructured mesh is represented as a graph $\mathcal{G} = (\mathcal{V}, \mathcal{E})$, where each node $i \in \mathcal{V}$ corresponds to a vertex of the finite-element mesh, and edges $(i,j) \in \mathcal{E}$ encode local connectivity between adjacent mesh vertices. The state variable is the vertex-wise velocity vector $\mathbf{u}_i(t) = [u_i(t),\, v_i(t)]^{\top} \in \mathbb{R}^{2}$, and the GNN maps the graph-valued state to the predicted temporal derivative $\dot{\mathbf{u}}(t)$. The temporal evolution of the flow field is modeled using a Neural ODE of the form
\begin{equation}
\frac{d\mathbf{u}(t)}{dt} = f_{\theta}\big(\mathbf{u}(t), \mathcal{G}\big),
\end{equation}
where $f_{\theta}$ is parameterized by a message-passing GNN and implemented as the right-hand side function within the \texttt{torchdiffeq} framework.

For each directed edge $(i,j)$, we construct geometric edge features
\begin{equation}
\mathbf{e}_{ij} =
\big[\Delta x_{ij},\, \Delta y_{ij},\, \|\Delta \mathbf{r}_{ij}\|\big]
\in \mathbb{R}^{3},
\end{equation}
where $\Delta \mathbf{r}_{ij} = \mathbf{p}_j - \mathbf{p}_i$ denotes the relative position between neighboring mesh vertices. These geometric edge attributes are embedded through an edge encoder multilayer perceptron (MLP), while the node features are embedded through a node encoder MLP into a latent space of dimension $H=128$. Message passing is performed using a mean aggregation operator. Specifically, at each propagation step, edge-wise messages are computed as
\begin{equation}
\mathbf{m}_{ij} =
\phi_{\theta}\!\left(
\left[\mathbf{h}_j,\, \mathbf{g}(\mathbf{e}_{ij})\right]
\right),
\qquad
\mathbf{m}_i = \frac{1}{|\mathcal{N}(i)|}
\sum_{j \in \mathcal{N}(i)} \mathbf{m}_{ij},
\end{equation}
followed by a node update
\begin{equation}
\mathbf{h}_i \leftarrow
\psi_{\theta}\!\left(
\left[\mathbf{h}_i,\, \mathbf{m}_i\right]
\right),
\end{equation}
where $\mathbf{g}(\cdot)$ denotes the edge encoder, and $\phi_{\theta}(\cdot)$ and $\psi_{\theta}(\cdot)$ correspond to the message and update MLPs, respectively. Two consecutive message-passing layers are applied, and a residual connection from the raw node inputs is introduced via a linear projection. Finally, a decoder MLP maps the latent node representations to the time derivative $\dot{\mathbf{u}}(t)$. This design allows the NODE to capture nonlinear interactions and spatial correlations on the unstructured mesh topology.

\paragraph{Training Details}
The GNN-based NODE is trained by minimizing a multi-step rollout loss on trajectory segments of length $L=3$. Given a start time index $t_0$, the initial condition is set to $\mathbf{u}(t_0)$, and the Neural ODE is integrated forward using the explicit fourth-order Runge--Kutta (RK4) method over $L$ rollout steps to obtain predictions $\hat{\mathbf{u}}(t_0+k)$ for $k=1, 2, \dots, L$. The training objective is the mean-squared error over the rollout segment,
\begin{equation}
\mathcal{L}(\theta)=\frac{1}{L}\sum_{k=1}^{L}
\left\|\hat{\mathbf{u}}(t_0+k)-\mathbf{u}(t_0+k)\right\|_2^2,
\end{equation}
where the norm is evaluated over all mesh vertices and both velocity components. At each training epoch, 80 random segments are sampled from the training interval. The dataset is normalized prior to training: the node features are standardized using the global mean and standard deviation computed over all timesteps and vertices, and the geometric edge attributes $\mathbf{e}_{ij}$ are similarly standardized using precomputed edge-wise statistics. Training was performed using the Adam optimizer with a learning rate of $10^{-3}$ on a single NVIDIA A100 GPU on ALCF Polaris for 2700 epochs, requiring approximately 3 hours in total. 

\subsection{Applying DEIM to Learned Dynamics}
\label{subsec:Applying DEIM to Learned Dynamics}
When applying DEIM to the trained NODE models, we employ a time-windowed DEIM approach. In this method, the right-hand side of each PDE is collected over a fixed time interval to form a snapshot matrix. The sampling matrix $P$ is then computed from these snapshots, after which the window is shifted forward by a stride $s$ to construct consecutive snapshot matrices. This procedure enables us to track the trajectories of representative sampling points during simulations and to evaluate the learned dynamics of the NODE models by comparing them with the trajectories obtained from the ground-truth data. For the two-dimensional vortex-merging problem, the window size and the number of DEIM sampling points are both set to 32. 

Formally, let $\mathbf{f}(t) \in \mathbb{R}^{n}$ denote the RHS snapshot used for DEIM, where we set
$\mathbf{f}(t) = \mathcal{R}(u(t))$ for the ground-truth solver, and $\mathbf{f}(t)=f_{\mathrm{NODE}}(u(t);\theta)$ for the NODE rollout.
Given a window size $W$ and stride $s$, we define the windowed snapshot matrix
\begin{equation}
\mathbf{F}_k = [\mathbf{f}(t_k),\,\mathbf{f}(t_k+1),\,\ldots,\,\mathbf{f}(t_k+W-1)] \in \mathbb{R}^{n\times W},
\qquad t_k = t_0 + ks.
\end{equation}
We compute a rank-$l$ basis $\Phi_k \in \mathbb{R}^{n\times l}$ from $\mathbf{F}_k$ via SVD and apply the standard greedy DEIM procedure to obtain interpolation indices
$S_k=\{p_{k,1},\ldots,p_{k,l}\}$.
We refer to the sequence $\{S_k\}_{k\ge 0}$ as the \emph{DEIM trajectory}.

\begin{figure}
  \begin{center}
      \includegraphics[width=\linewidth]{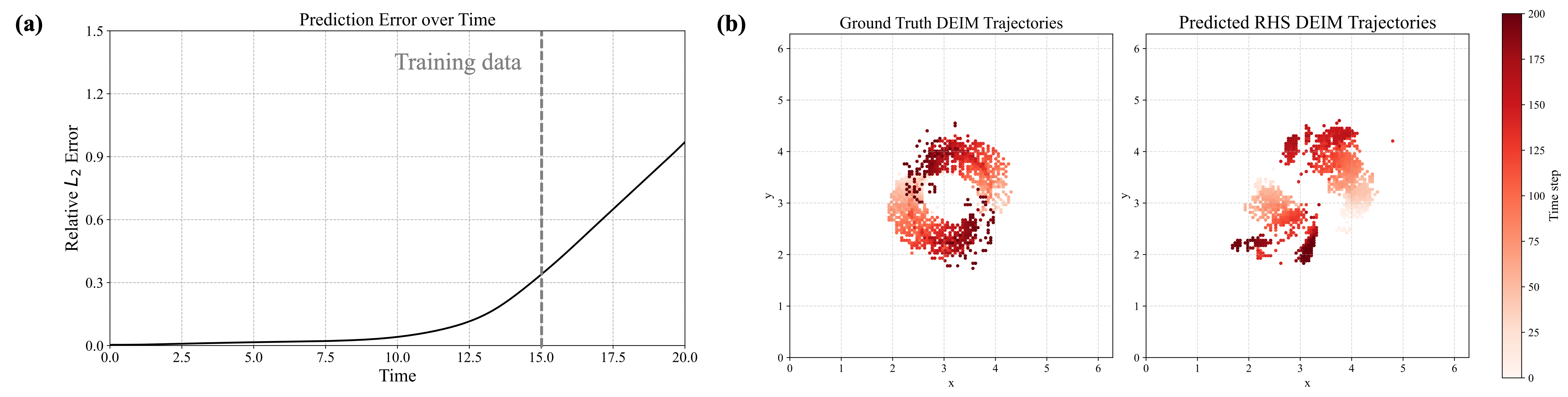}
  \end{center}
  \caption{For the case with initial horizontal vortices: (a) Relative $L_2$-norm error at each timestep during rollout; (b) trajectories of sampling points from the ground truth and the NODE prediction.}
  \label{fig:sampling_VM}
\end{figure}

Figure \ref{fig:sampling_VM}(b) shows the trajectories of sampling points obtained by the DEIM from the ground truth and the NODE prediction for the case with initially symmetric horizontal vortices shown in Fig. \ref{fig:vortexmerging_inital_final}(a). The color bar indicates time progression, with lighter (near-white) colors corresponding to earlier time steps and darker red colors indicating later times. As the two vortices rotate counterclockwise and merge, the DEIM-selected sampling points exhibit a corresponding counterclockwise rotation. Due to the symmetry of the vortex pair, the sampling points predominantly circulate around the two vortex cores. During the first half of the simulation, the sampling-point trajectories extracted from the NODE prediction closely follow those from the ground truth. However, at later times, the NODE-based trajectories deviate from the circular motion, repeatedly revisiting specific points. This deviation is consistent with the evolution of the relative $L_2$-norm error shown in Fig. \ref{fig:sampling_VM}(a), where the error begins to grow after approximately $t = 10.0$. This behavior suggests that the NODE fails to capture the long-term vortex interaction dynamics, leading to a collapse of the effective sampling manifold.

\begin{figure}
  \begin{center}
      \includegraphics[width=\linewidth]{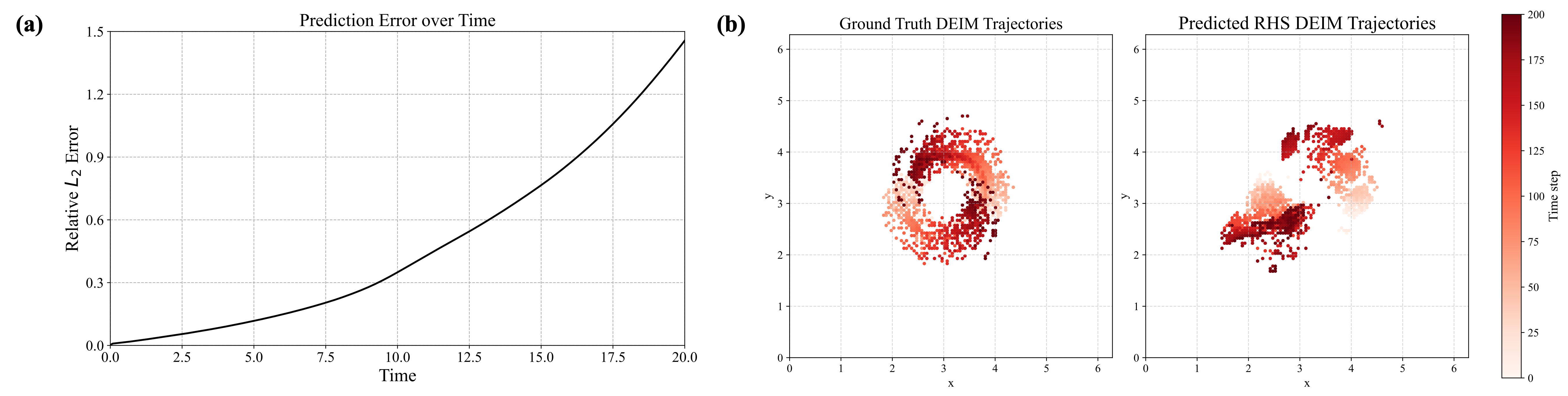}
  \end{center}
  \caption{For the case with initial asymmetric horizontal vortices: (a) Relative $L_2$-norm error at each timestep during rollout; (b) trajectories of sampling points from the ground truth and the NODE prediction.}
  \label{fig:sampling_VM_asymmetric}
\end{figure}

Figure \ref{fig:sampling_VM_asymmetric} follows the same layout as Fig. \ref{fig:sampling_VM}, but corresponds to the case of initially asymmetric horizontal vortices shown in Fig. \ref{fig:vortexmerging_inital_final}(b). This configuration represents an extrapolation scenario, as the model was not exposed to such snapshots during training. As shown in Fig. \ref{fig:sampling_VM_asymmetric}(b), the trajectories of the DEIM-selected sampling points initially exhibit a partially circular pattern during the early stages of the simulation. However, this regular structure rapidly breaks down, and the trajectories begin to oscillate irregularly. This behavior is consistent with the evolution of the relative $L_2$-norm error in Fig. \ref{fig:sampling_VM_asymmetric}(a), where the error increases sharply shortly after the start of the simulation.

\begin{figure}
  \begin{center}
      \includegraphics[width=\linewidth]{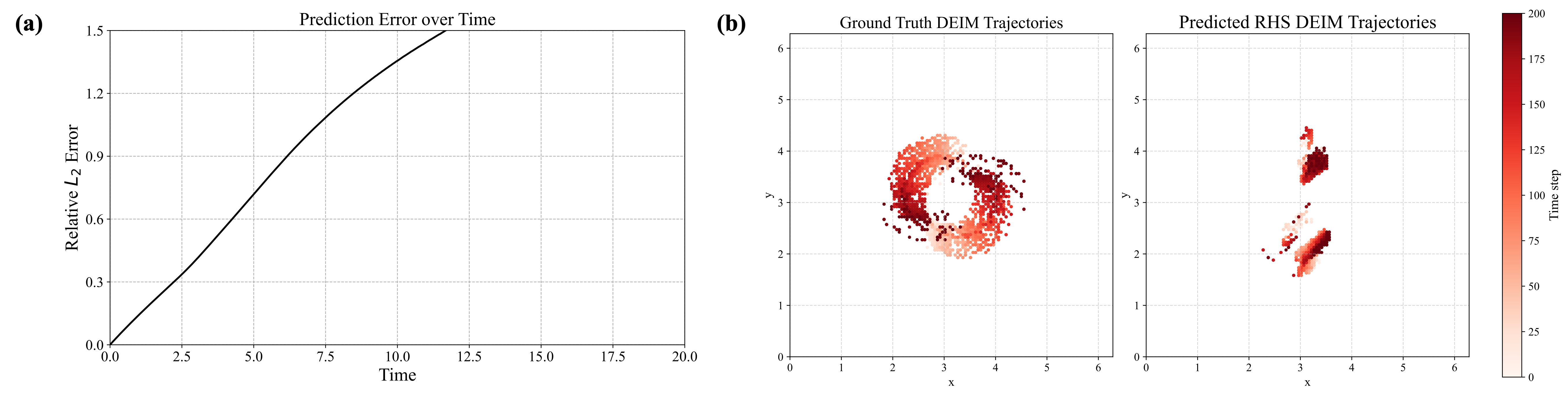}
  \end{center}
  \caption{For the case with initial vertical vortices: (a) Relative $L_2$-norm error at each timestep during rollout; (b) trajectories of sampling points from the ground truth and the NODE prediction.}
  \label{fig:sampling_VM_vertical}
\end{figure}

Figure \ref{fig:sampling_VM_vertical} corresponds to the case with initially vertical vortices shown in Fig. \ref{fig:vortexmerging_inital_final}(c). In this configuration, the trajectories of the DEIM-based sampling points extracted from the NODE prediction do not exhibit any coherent circular structure, even at the early stages of the simulation. Instead of maintaining rotational motion, the sampling points rapidly collapse into a limited set of spatial regions. Compared to the previously considered case with initially asymmetric vortices, this concentration of sampling points occurs significantly earlier, indicating a faster breakdown of the learned dynamical structure. This behavior is also reflected in the relative $L_2$-norm evolution in Fig. \ref{fig:sampling_VM_vertical}(a), where the prediction error for the initially vertical vortex configuration increases more rapidly than in the asymmetric vortex case.

\begin{figure}
  \begin{center}
      \includegraphics[width=0.7\linewidth]{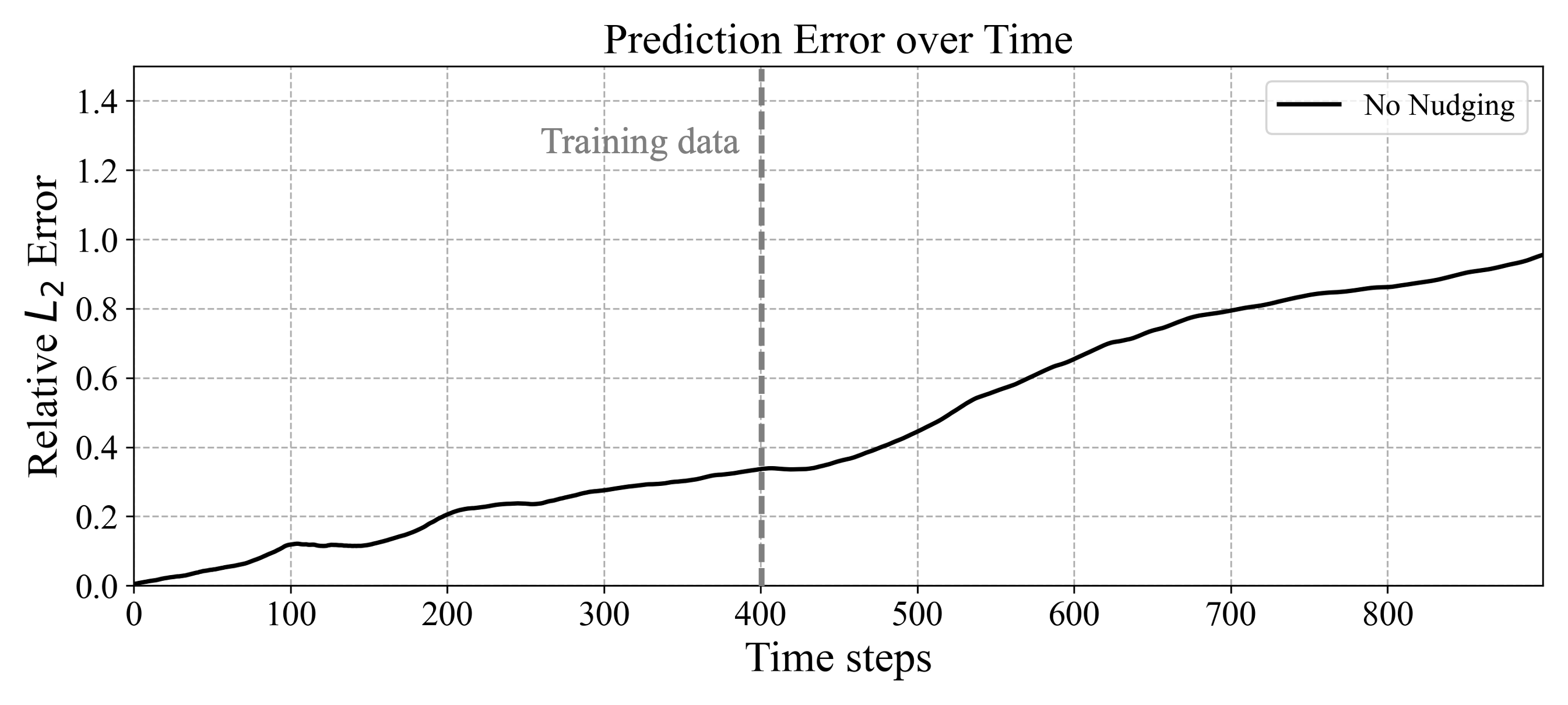}
  \end{center}
  \caption{For the two-dimensional BFS case: (a) the relative $L_2$ error from the NODE prediction during rollout; (b) the comparison of contours of streamwise velocity from the ground truth and NODE prediction at different timesteps}
  \label{fig:BFS_error}
\end{figure}

\begin{figure}
  \begin{center}
      \includegraphics[width=0.7\linewidth]{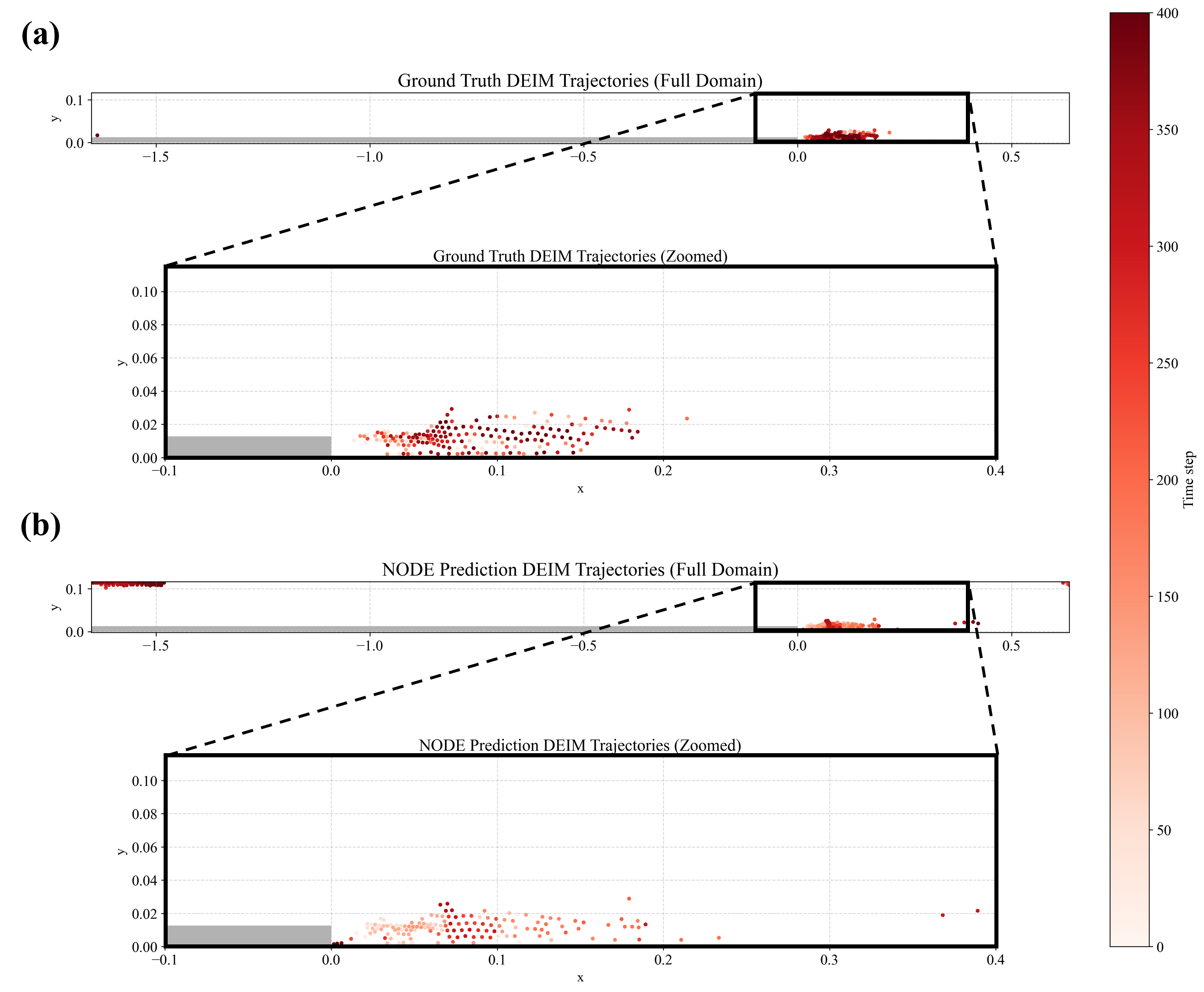}
  \end{center}
  \caption{For the two-dimensional BFS case: (a) trajectory of DEIM-sampling points from the ground truth; (b) trajectory of DEIM-sampling points from the NODE prediction.}
  \label{fig:BFS_sampling}
\end{figure}

For the two-dimensional BFS case, a window size of 24 is used to extract sampling points via DEIM. When applying DEIM to the windowed snapshots, the streamwise and wall-normal velocity components are treated as separate state variables. Specifically, instead of forming a snapshot matrix of size $[N,T]$ for a scalar field, we construct an augmented snapshot matrix of size $[2N,T]$ by stacking the two velocity components. This formulation allows DEIM to identify dynamically important sampling locations independently for each velocity component. Figure \ref{fig:BFS_sampling} compares the trajectories of DEIM-sampled points extracted from the ground-truth solution and the NODE prediction for the two-dimensional backward-facing step (BFS) case, using a DEIM window size of 24. As shown in Fig. \ref{fig:BFS_sampling}(a), most DEIM-sampled points from the ground truth are concentrated downstream of the step, where vortices shed from the step edge are convected. Specifically, the sampling points are initially located near the step edge, where vortices form due to flow separation, and subsequently move along with the evolving vortex structures. Once a vortex dissipates due to viscous effects, new sampling points emerge near newly formed vortices at the step edge and again follow their trajectories. In contrast, Fig. \ref{fig:BFS_sampling}(b) shows that the DEIM-sampled points obtained from the NODE prediction initially remain near the step edge and exhibit trajectories similar to those of the ground truth during the early stages of the rollout. However, as prediction errors accumulate over time, the sampling points gradually drift away from the vortex trajectories and instead concentrate near the channel inlet. This behavior is consistent with the relative $L_2$-norm error evolution shown in Fig. \ref{fig:BFS_error}, indicating that the collapse of DEIM-sampled point trajectories serves as a diagnostic indicator of NODE prediction failure.

\subsection{DEIM-based Data Assimilation for Neural ODE}
In this section, we leverage diagnostic information extracted from trained NODEs via DEIM-based sampling within a data assimilation framework to improve predictive performance. In particular, we adopt a nudging-based data assimilation approach. Nudging stabilizes dynamical system predictions by gently relaxing the model state toward available observations \cite{wu2025interpolated, plogmann2026sparse}. For a standard PDE system, this is achieved by introducing a correction term proportional to the discrepancy between the observed field and the model state.

\begin{equation}
\label{eqn:pde_nudging}
\frac{\partial u}{\partial t} = \mathcal{R}(u) + \gamma (u_{obs}-u)
\end{equation}

where $u$ and $u_{\mathrm{obs}}$ denote the predicted and observed states, respectively, $\mathcal{R}$ represents the PDE operator, and $\gamma$ is the nudging parameter. The discrepancy between the observed and predicted states acts as a source term that continuously steers the system toward the observations. We extend this nudging formulation to our pre-trained Neural ODE model. In particular, nudging source terms are applied only at the DEIM-selected spatial locations identified in the previous section.

\begin{equation}
\frac{\partial u}{\partial t}
=
f_{\mathrm{NODE}}(u;\theta)
+
\gamma \, \mathbf{P}_S^\top
\bigl(\mathbf{y}_{\mathrm{obs}} - \mathbf{P}_S u\bigr),
\qquad \mathbf{y}_{\mathrm{obs}} = \mathbf{P}_S u_{\mathrm{obs}}.
\end{equation}

, where $f_{\mathrm{NODE}}(u;\theta)$ denotes the NODE approximation of the underlying PDE operator, and $\mathbf{P}_S$ is a projection (selection) operator that extracts the state variables at the DEIM-selected spatial locations $S$. The selection set $S$ is constructed online using the same windowed-DEIM procedure described in Sec.~3.2, applied to a sliding window of the NODE RHS snapshots. $\mathbf{P}_S^\top$ injects the nudging corrections back into the full state space, resulting in a sparse forcing term that acts only at the DEIM-selected degrees of freedom. 

However, nudging exclusively at DEIM-selected points may be insufficient, as these points typically constitute only a small fraction of the full computational domain. For instance, in the two-dimensional vortex-merging problem considered in the previous section, DEIM sampling based on 32 prior snapshots selects approximately 0.2\% of the total spatial points. With such sparse coverage, the resulting nudging signal is often too localized to effectively steer the global dynamics toward the ground-truth trajectory. To address this limitation, we enrich the nudging locations by expanding the DEIM-sampled point set using kernel density estimation (KDE). Specifically, we treat the DEIM-selected locations as samples from an underlying spatial importance distribution and construct a continuous density estimate via KDE. Additional nudging points are then drawn from high-density regions of this estimated distribution, thereby preserving the spatial structure identified by DEIM while increasing the overall nudging coverage. In practice, when 32 snapshots are used to identify DEIM points, this KDE-based expansion increases the number of nudging locations to 256 points, corresponding to approximately 1.5\% of the total grid points in the vortex-merging problem. This enriched nudging set provides a more effective balance between sparsity and global stabilization.

Concretely, given DEIM-selected coordinates $\{\mathbf{x}_i\}_{i=1}^{M}$, we form a KDE
\begin{equation}
\hat{p}(\mathbf{x})=\frac{1}{M h^d}\sum_{i=1}^{M}K\!\left(\frac{\|\mathbf{x}-\mathbf{x}_i\|}{h}\right),
\end{equation}
where $d$ is the spatial dimension, $h$ is the bandwidth, and $K(\cdot)$ is a Gaussian kernel.
We then select additional nudging locations by sampling (or ranking grid points) according to $\hat{p}(\mathbf{x})$ and taking the top $M_{\mathrm{add}}$ locations, producing an enriched set $S^{\mathrm{KDE}}$.

\begin{figure}
  \begin{center}
      \includegraphics[width=\linewidth]{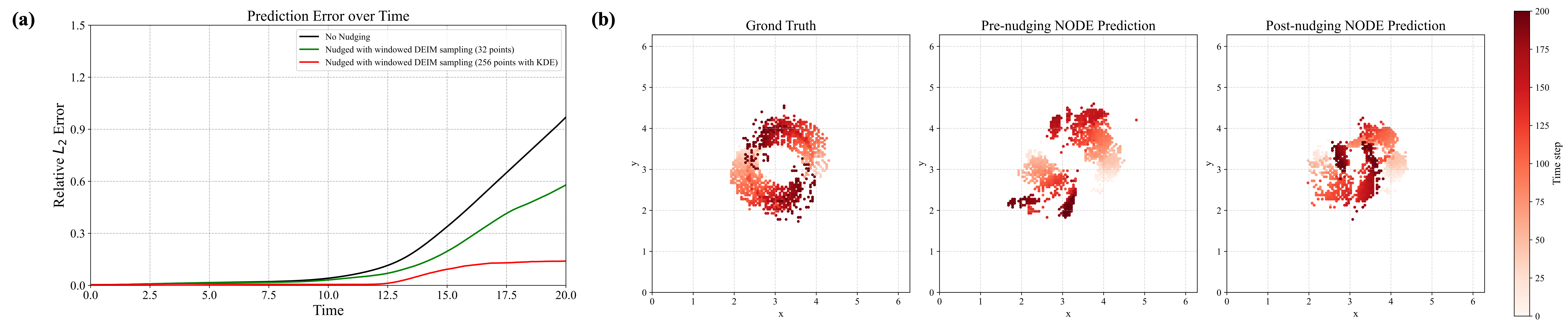}
  \end{center}
\caption{For the case with initially horizontal vortices:
(a) Relative $L_2$ error during rollout for the raw NODE prediction, nudging at DEIM-selected locations only (32 points),
and nudging at DEIM locations enriched via KDE (32 DEIM points expanded to a total of 256 locations);
(b) DEIM sampling-point trajectories for the ground truth, the raw NODE rollout, and the post-nudging rollout
(DEIM+KDE), shown with color indicating time progression.}
  \label{fig:sampling_VM_nudging}
\end{figure}

Figure \ref{fig:sampling_VM_nudging}(a) compares the relative $L_2$-norm error at each timestep during rollouts from the NODE prediction before and after applying DEIM-based nudging with KDE enrichment. While nudging using only the DEIM-selected points yields only marginal improvements in prediction accuracy and stability, incorporating additional nudging locations populated via KDE leads to a substantial reduction in error and a marked stabilization of the dynamics. By applying nudging at both the DEIM-selected points and the KDE-enriched locations—where the system dynamics are most strongly represented—the NODE prediction more accurately follows the underlying evolution of the flow. This improvement is further illustrated in Fig. \ref{fig:sampling_VM_nudging}(b), which compares the trajectories of the DEIM-selected points for the ground-truth solution, the unnudged NODE prediction, and the NODE prediction with DEIM-based nudging and KDE enrichment. The nudged NODE trajectory exhibits a significantly more coherent and stable circular structure, closely resembling the ground-truth behavior. These results demonstrate that nudging at dynamically representative sampling locations enables the NODE to more faithfully capture the vortex interaction dynamics.

\begin{figure}
  \begin{center}
      \includegraphics[width=\linewidth]{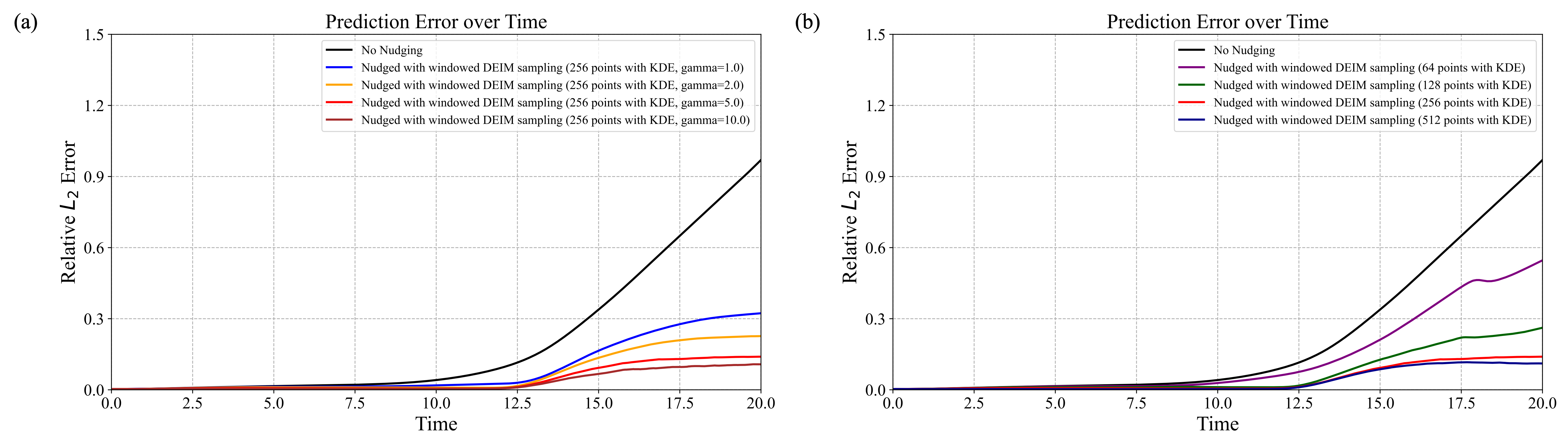}
  \end{center}
  \caption{For the case with initially horizontal vortices: Comparison of relative $L_2$ errors for DEIM-based nudging under different settings: (a) nudging gain values and (b) number of sampling points}
  \label{fig:vm_error_parameter_compare}
\end{figure}

Figure~\ref{fig:vm_error_parameter_compare} summarizes a parameter sweep of the proposed DEIM-based nudging strategy for the case with initially horizontal vortices. In Fig.~\ref{fig:vm_error_parameter_compare}(a), we fix the KDE-enriched DEIM nudging set to 256 locations and vary the nudging gain $\gamma$. While the raw NODE rollout exhibits rapid error growth in the late-time regime, increasing $\gamma$ consistently suppresses this growth and yields a markedly lower final relative $L_2$ error, indicating that stronger relaxation toward sparse observations improves long-horizon stability. In Fig.~\ref{fig:vm_error_parameter_compare}(b), we instead hold the nudging gain fixed and vary the number of nudging locations generated by KDE (64--512 points). As the nudging budget increases, the rollout becomes progressively more stable and accurate, highlighting that sufficient spatial coverage is necessary for sparse corrections to influence the global vortex interaction dynamics. At the same time, the improvement begins to saturate for larger budgets, suggesting a practical trade-off between accuracy and the cost of increasing the number of observation updates. Collectively, these sweeps motivate the choice of $\gamma$ and the nudging budget used in the subsequent vortex-merging experiments.

\begin{figure}
  \begin{center}
      \includegraphics[width=\linewidth]{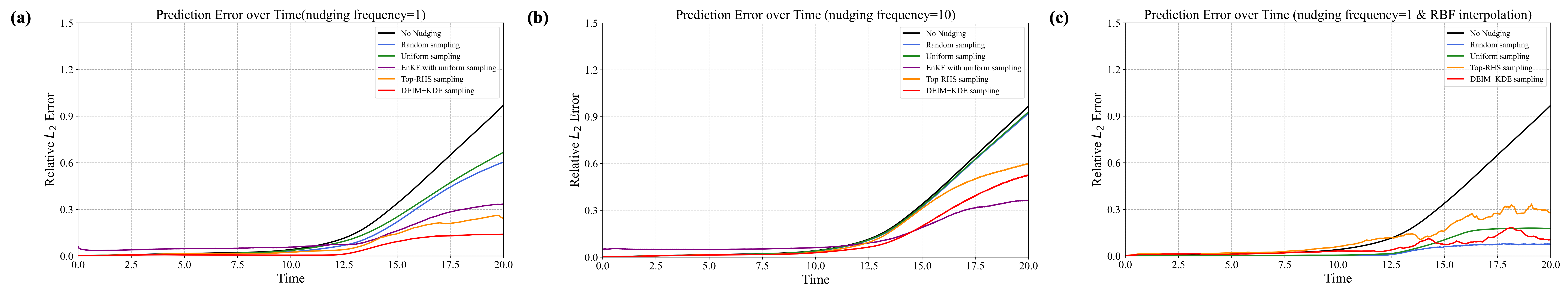}
  \end{center}
  \caption{Comparison of relative $L_2$ errors from NODE predictions under different sparse data assimilation baselines
and sampling policies: (a) without RBF interpolation, (b) without RBF interpolation and nudging every 10 timesteps, and (c) with RBF interpolation.}
  \label{fig:VM_error_nudging_compare}
\end{figure}

Figure~\ref{fig:VM_error_nudging_compare} compares the relative $L_2$ errors of NODE rollouts under different sparse nudging strategies, including random sampling, uniformly distributed sampling, ensemble Kalman filter (EnKF) with uniform sampling, top-RHS sampling, and the proposed KDE-enriched DEIM sampling. To ensure a fair comparison, the total number of nudging locations is held fixed across all strategies. In addition to injecting pointwise corrections only at the selected locations, we also consider an optional spatial extension operator based on radial basis function (RBF) interpolation, which reconstructs a smooth correction field from the sparse discrepancies and distributes the nudging signal over the full domain. This allows us to isolate how the effectiveness of a sampling policy changes when sparse observation errors are propagated beyond the directly nudged degrees of freedom. When nudging is applied at every timestep without RBF interpolation (Fig.~\ref{fig:VM_error_nudging_compare}(a)), the DEIM+KDE strategy yields the most pronounced long-term stabilization, producing the smallest error growth among all methods, while random and uniform sampling provide only modest improvements over the unnudged rollout. In this setting, the top-RHS heuristic and the EnKF baseline reduce the error relative to random/uniform nudging, but remain less effective than allocating the nudging budget to DEIM-identified, dynamically representative regions. When the nudging frequency is reduced to every 10 timesteps (Fig.~\ref{fig:VM_error_nudging_compare}(b)), the overall benefit of sparse nudging decreases—most notably for random and uniform sampling—and the performance gap between strategies narrows, indicating that temporally sparse corrections are insufficient unless the update mechanism is sufficiently informative. Finally, introducing RBF interpolation (Fig.~\ref{fig:VM_error_nudging_compare}(c)) substantially improves stability across all sampling strategies by spatially spreading the sparse discrepancy information, and the ranking becomes more sensitive to spatial coverage rather than purely structure-aware point selection. Overall, these results highlight that the effectiveness of sparse data assimilation depends jointly on (i) where observations are placed (sampling policy), (ii) how often corrections are injected (nudging frequency), and (iii) how localized discrepancies are propagated to the full state (interpolation/extension operator).

\begin{figure}
  \begin{center}
      \includegraphics[width=\linewidth]{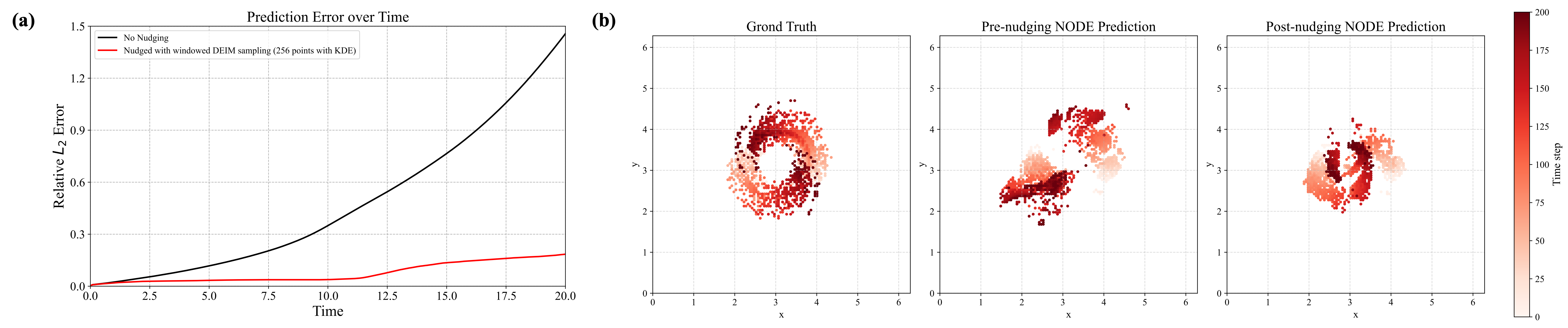}
  \end{center}
  \caption{For the case with initial asymmetric horizontal vortices: (a) Relative $L_2$-norm error at each timestep during rollout for the baseline NODE prediction and the nudged NODE prediction; (b) trajectories of DEIM sampling points comparing the ground truth, the baseline NODE prediction, and the nudged NODE prediction.}
  \label{fig:sampling_VM_asymmetric_nudging}
\end{figure}

\begin{figure}
  \begin{center}
      \includegraphics[width=\linewidth]{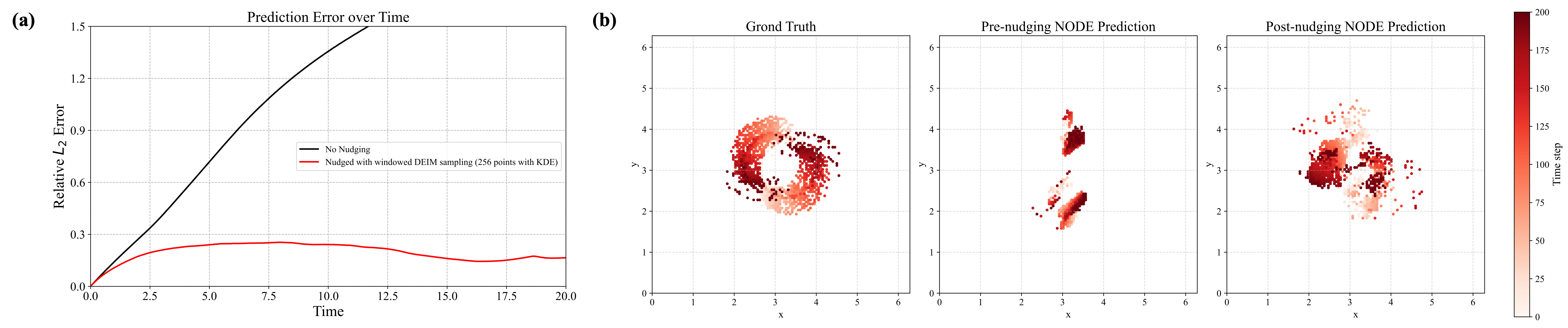}
  \end{center}
  \caption{For the case with initial vertical vortices: (a) Relative $L_2$-norm error at each timestep during rollout for the baseline NODE prediction and the nudged NODE prediction; (b) trajectories of DEIM sampling points comparing the ground truth, the baseline NODE prediction, and the nudged NODE prediction.}
  \label{fig:sampling_VM_vertical_nudging}
\end{figure}

The relative $L_2$-norm error plot in Fig. \ref{fig:sampling_VM_asymmetric_nudging}(a) demonstrates that DEIM-based nudging significantly improves both the stability and accuracy of NODE predictions in the extrapolation case with initially asymmetric vortices. Figure \ref{fig:sampling_VM_asymmetric_nudging}(b) shows that the trajectories of DEIM-sampled points from the nudged NODE prediction more consistently follow the circular motion of the two vortices, whereas the trajectories from the raw NODE prediction deviate from this coherent structure. This improved structural consistency indicates enhanced predictability of the underlying system dynamics and aligns with the observed reduction in relative $L_2$ error. A similar trend is observed in the more challenging extrapolation case with initially vertical vortices. The raw NODE prediction rapidly loses the coherent rotational structure and exhibits collapsing sampling trajectories. In contrast, the DEIM-nudged prediction (Fig. \ref{fig:sampling_VM_vertical_nudging}(b)) maintains dynamically consistent sampling paths that track the two rotating vortices. Notably, even in this test case—where the unnudged NODE performs poorly—DEIM-based nudging markedly improves both predictive stability and accuracy.

% \begin{figure}
%   \begin{center}
%       \includegraphics[width=0.7\linewidth]{Figure/error_relativel2norm_over_time_sampling_methods_RBF_compare.pdf}
%   \end{center}
%   \caption{Comparison of $L_2$ norm errors from the NODE prediction with different nudging strategies.}
%   \label{fig:error_nudging_compare}
% \end{figure}

\begin{figure}
  \begin{center}
      \includegraphics[width=0.7\linewidth]{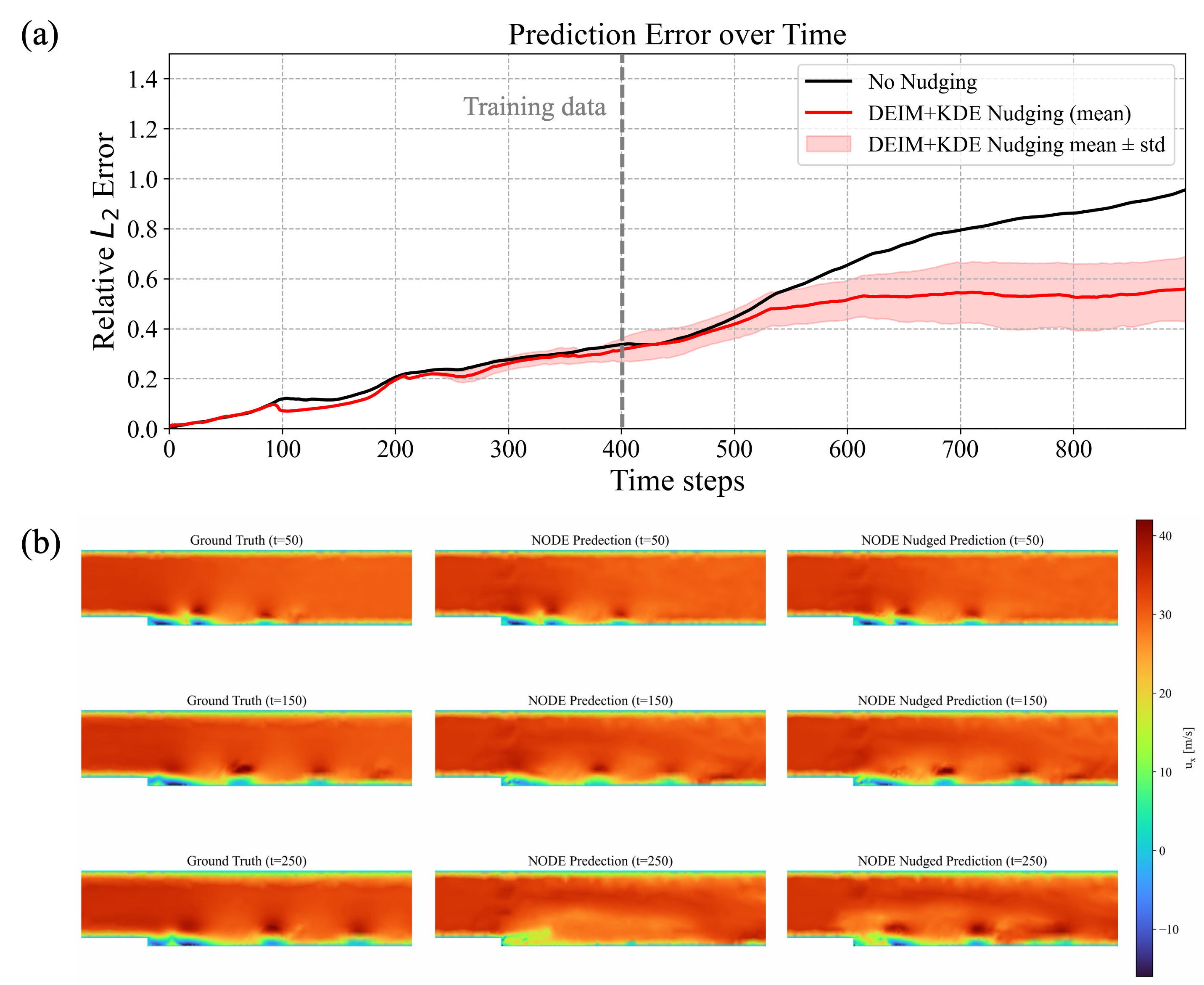}
  \end{center}
\caption{For the two-dimensional BFS case: (a) relative $L_2$ error during rollout. For DEIM+KDE nudging, the solid red curve denotes the ensemble mean over 10 repeated runs (run-to-run variability induced by the non-smooth DEIM/top-$k$ point-selection operator), and the shaded region indicates $\pm$ one standard deviation. The dashed vertical line marks the end of the training horizon. (b) Comparison of streamwise-velocity contours at selected time instances for the ground truth (left), raw NODE rollout without nudging (middle), and DEIM+KDE-nudged NODE rollout (right), shown for a representative ensemble member.}
\label{fig:BFS_error_nudged_uq}
\end{figure}

\begin{figure}
  \begin{center}
      \includegraphics[width=\linewidth]{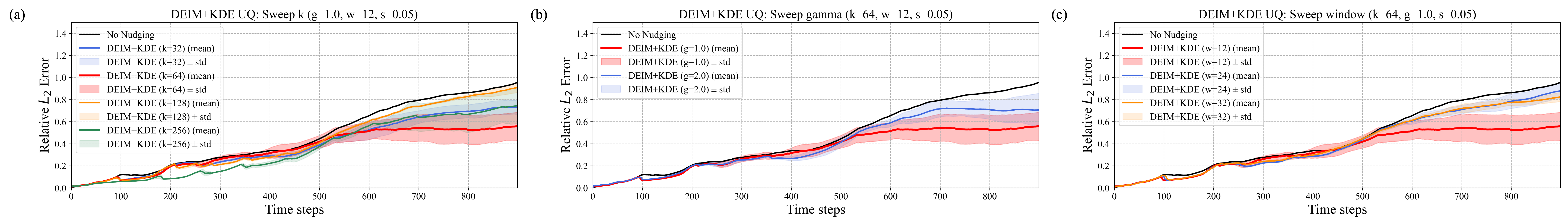}
  \end{center}
  \caption{Comparison of relative $L_2$ errors for DEIM-based nudging under different settings: (a) number of sampling points, (b) nudging gain values, and (c) DEIM window sizes. Solid curves denote ensemble means and shaded regions indicate $\pm$ one standard deviation over 10 realizations.}
  \label{fig:BFS_sampling_nudging_sweep}
\end{figure}

For the two-dimensional BFS case, we reduce the DEIM window size from 24 in the diagnostic analysis (Sec. \ref{subsec:Applying DEIM to Learned Dynamics}) to 12 in the data-assimilation experiments, with 12 DEIM sampling points selected accordingly. This shorter window better captures the rapidly convecting and intermittently forming structures near the step, as discussed later in this section and illustrated in Fig. \ref{fig:BFS_sampling_nudging_sweep}. These points are then expanded to 64 locations (approximately $2.0\%$ of the total grid points) using KDE. In practice, however, the BFS flow exhibits strong advection and unsteady separation dynamics, and long-horizon NODE rollouts can be highly sensitive to small perturbations. In our framework, such perturbations arise naturally because the nudging operator is constructed from a hard (discrete) point-selection pipeline---including the greedy DEIM index selection and a KDE-based enrichment step implemented via density ranking/top-$k$ selection---which induces a non-smooth dependence of the selected set on the snapshot window/state (so that small numerical changes can switch the selected indices). As a result, even when the nudging budget and all other hyperparameters are fixed (and the NODE parameters are kept unchanged), repeated runs can yield noticeably different late-time rollout trajectories due to cumulative error amplification (often manifesting as phase drift in the shed-vortex dynamics). To account for this sampling-induced sensitivity, we repeat the DEIM+KDE nudging rollout over multiple runs and report the ensemble mean of the relative $L_2$ error together with $\pm$ one standard deviation, as shown by the shaded band in Fig.~\ref{fig:BFS_error_nudged_uq}(a).

\begin{figure}
  \begin{center}
      \includegraphics[width=0.7\linewidth]{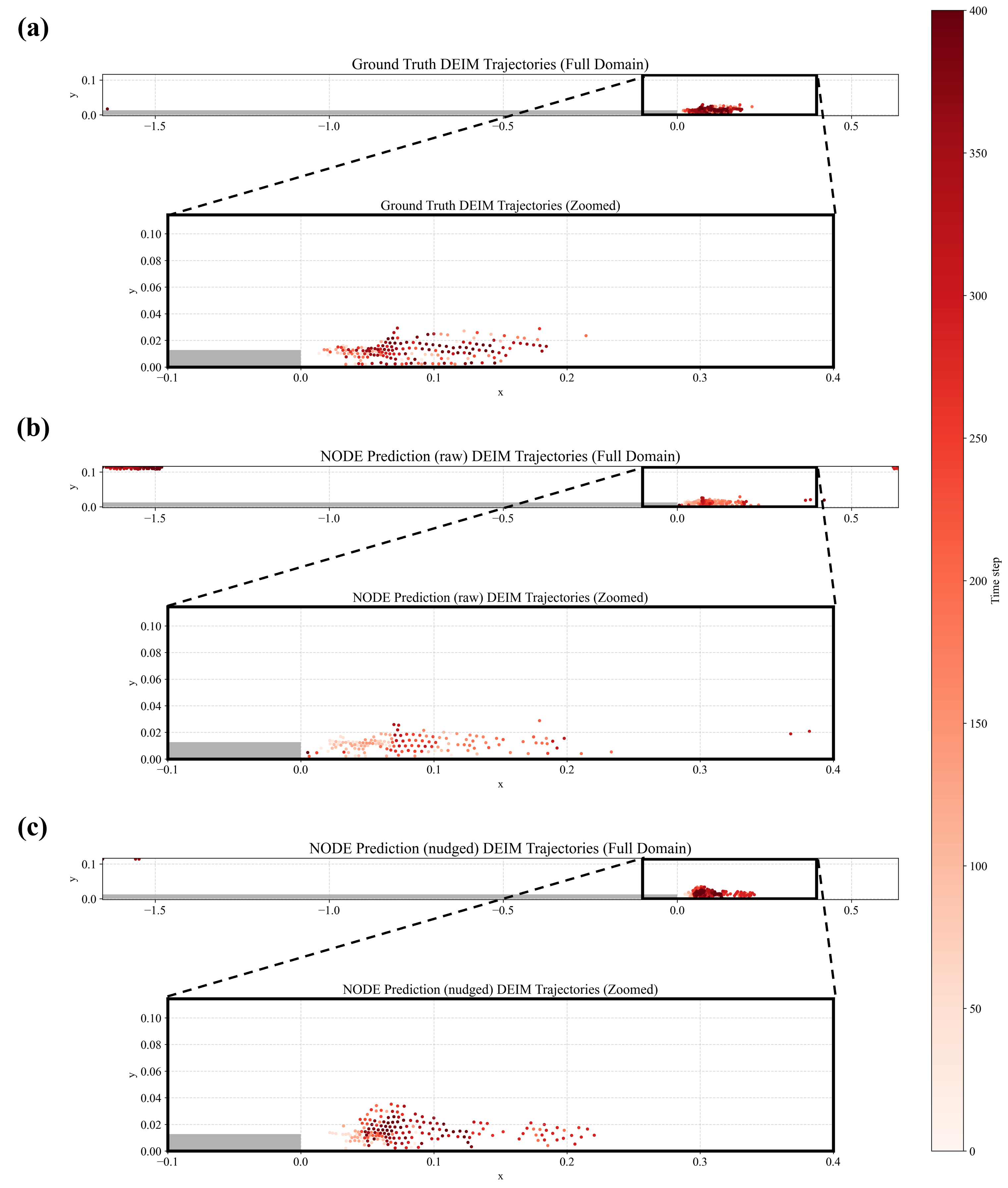}
  \end{center}
  \caption{For the two-dimensional BFS case: trajectories of DEIM-sampling points, with color indicating time progression. For each case, the top panel shows the full domain and the bottom panel shows a zoomed view near the step/separation region: (a) ground truth, (b) raw NODE prediction, and (c) NODE prediction with DEIM+KDE-based nudging.}
  \label{fig:BFS_sampling_nudging}
\end{figure}

Figure~\ref{fig:BFS_error_nudged_uq}(b) visualizes streamwise-velocity contours from the ground truth (left column), the raw, unnudged NODE prediction (middle column), and the DEIM+KDE-nudged NODE prediction (right column) at selected time instances using a representative ensemble member. At early times, all three fields remain close, but as the rollout proceeds the raw NODE prediction exhibits increasingly apparent phase and amplitude errors in the shed-vortex dynamics and downstream convection patterns. In contrast, the DEIM+KDE-nudged rollout better preserves the dominant downstream structures associated with separation, vortex shedding, and convective transport, delaying the onset of large-scale drift relative to the raw NODE. The remaining differences at later times are primarily associated with small phase offsets and localized amplitude mismatches, which are strongly amplified by the advection-dominated, unsteady nature of the BFS flow and contribute to the ensemble spread observed in Fig.~\ref{fig:BFS_error_nudged_uq}(a). This improvement is also reflected in the sampling trajectories shown in Fig.~\ref{fig:BFS_sampling_nudging}. While the sampling points from the raw NODE prediction collapse and concentrate near the channel inlet in the full-domain view (and consequently lose alignment with the downstream separated structures in the zoomed view), the nudged NODE prediction maintains sampling locations primarily downstream of the step, where vortices are shed and convected. This indicates improved tracking of the dominant flow dynamics.

Interestingly, we observe that the run-to-run variability induced by the discretely selected nudging operator is strongly case-dependent: for the vortex-merging case, we found negligible spread across repeated runs under the same hard-selection and KDE enrichment pipeline, whereas the BFS case exhibits a pronounced late-time ensemble spread. This suggests that the sensitivity of long-horizon NODE rollouts to sparse, discretely selected nudging configurations can depend substantially on the underlying dynamics, with advection-dominated separated flows being more prone to cumulative error amplification and phase drift than coherent vortex-merging dynamics.

\begin{figure}
  \begin{center}
      \includegraphics[width=1.0\linewidth]{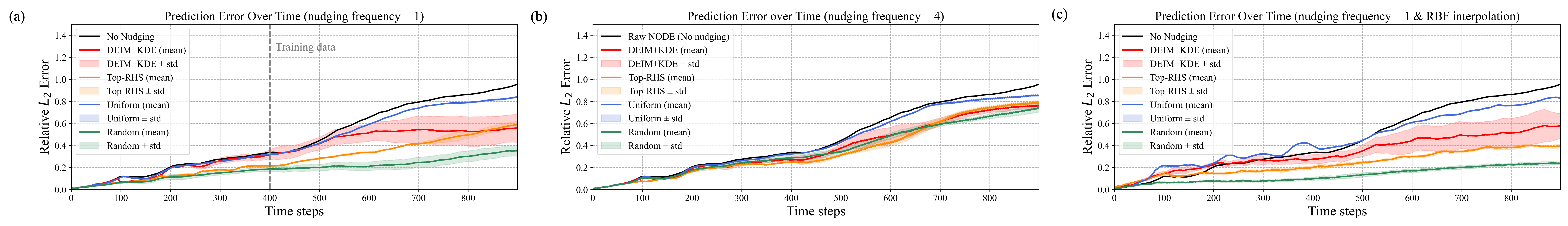}
  \end{center}
  \caption{Comparison of relative $L_2$ errors from the BFS NODE rollout under different sparse nudging strategies, reporting ensemble mean (solid) and $\pm$ one standard deviation (shaded) across multiple realizations of the sampling-point set: (a) Nudging applied at every timestep without RBF interpolation (the dashed vertical line marks the end of the training horizon). (b) Nudging applied every four timesteps without RBF interpolation. (c) Nudging applied at every timestep with RBF interpolation}

\label{fig:BFS_nudging_compare}
\end{figure}

Figure~\ref{fig:BFS_sampling_nudging_sweep} further summarizes a parameter sweep of the DEIM+KDE nudging configuration for the BFS case under the same ensemble protocol. Here, we denote by $k$ the total number of nudging locations after KDE enrichment, by $w$ the DEIM window size used to construct the sampling operator, and by $\gamma$ the nudging gain (with the KDE bandwidth fixed in this sweep). In Fig.~\ref{fig:BFS_sampling_nudging_sweep}(a), increasing the nudging budget does not lead to a strictly monotonic improvement: a moderate budget (e.g., $k=64$) yields the most consistent late-time error reduction, whereas larger budgets ($k\ge 128$) can provide diminishing returns or even degrade performance, suggesting that overly dense sparse forcing may interfere with the advection-dominated dynamics. In Fig.~\ref{fig:BFS_sampling_nudging_sweep}(b), increasing the nudging gain from $\gamma=1.0$ to $\gamma=2.0$ increases both the mean late-time error and the ensemble spread, indicating that overly aggressive relaxation can amplify phase errors in the separated shear-layer dynamics. Finally, Fig.~\ref{fig:BFS_sampling_nudging_sweep}(c) shows that shorter DEIM windows (e.g., $w=12$) outperform longer windows ($w\ge 24$), consistent with the fact that in BFS the most informative sampling locations are tied to rapidly convecting and intermittently forming structures, which can be blurred by longer time windows. Guided by these sweeps, we use the best-performing configuration as the default setting in the subsequent BFS comparisons across different sampling strategies (Fig.~\ref{fig:BFS_nudging_compare}).

Figure~\ref{fig:BFS_nudging_compare}(a) compares long-horizon rollout errors for the BFS Neural ODE under a fixed nudging budget using four different strategies for selecting nudging locations—(i) DEIM points enriched by KDE sampling (DEIM+KDE), (ii) points selected from the largest instantaneous RHS magnitude (Top-RHS), (iii) uniformly distributed points (Uniform), and (iv) randomly selected points (Random)—together with the unnudged baseline. Similar to the KDE-enriched DEIM sampling, the observation operator of the Top-RHS sampling is constructed through a discrete, non-smooth point-selection pipeline. As a result, small numerical differences can change the selected indices and lead to differences in the late-time trajectory. Moreover, for Random sampling the point set is explicitly resampled across realizations. We therefore report the ensemble mean of the relative $L_2$ error together with $\pm$ one standard deviation across multiple realizations (shaded bands), with the dashed vertical line indicating the end of the training horizon. Overall, all nudging strategies suppress post-training error growth relative to the raw NODE rollout, but their effectiveness differs markedly. Random nudging achieves the lowest mean error over most of the extrapolative regime, while Top-RHS provides strong early improvements immediately after the training cutoff and remains competitive at later times. Notably, Top-RHS can outperform DEIM+KDE in this BFS configuration because it selects nudging locations based on the \emph{instantaneous} magnitude of the NODE-predicted RHS, thereby adapting to the current (rapidly convecting) shear-layer and shed-vortex structures that dominate the error growth. In contrast, windowed-DEIM is constructed from a \emph{time window} of prior RHS snapshots, and thus can emphasize lagged or recurrent structures near the separation region rather than tracking the instantaneous downstream advection, reducing its effectiveness in strongly advection-dominated regimes. This regime dependence is consistent with our vortex-merging results, where the dominant dynamics are organized around a small number of coherent, slowly evolving structures and the windowed-DEIM sampling more reliably captures dynamically representative regions, making DEIM+KDE comparatively more effective. Interestingly, Uniform sampling performs substantially worse than Random and remains close to the unnudged baseline. A plausible explanation is that the BFS dataset is defined on a strongly nonuniform unstructured mesh (with refined resolution near the step and in the downstream recirculation/shear-layer region): Random sampling on mesh vertices naturally allocates a larger fraction of the nudging budget to these dynamically active, highly resolved regions, whereas a uniformly distributed selection in physical space spreads points more evenly across dynamically less informative areas, reducing the effective correction strength where it matters most.

Figure~\ref{fig:BFS_nudging_compare}(b) repeats the comparison when nudging is applied every four timesteps. As expected, reducing the update frequency weakens the stabilizing effect of sparse corrections: mean errors increase for all strategies and the separation between methods becomes smaller, with uncertainty bands overlapping more substantially in the extrapolative portion of the rollout. This indicates that, for the BFS flow, model errors can accumulate rapidly between correction events, so less frequent nudging provides limited opportunity to suppress the phase/amplitude drift associated with vortex shedding and convective transport. Nevertheless, even at this reduced update rate, nudging remains beneficial compared with the unnudged baseline, highlighting a practical trade-off between assimilation cost (frequency of observation injection) and long-horizon accuracy/stability.

Finally, Figure~\ref{fig:BFS_nudging_compare}(c) shows the corresponding results when nudging is applied at every timestep and the sparse discrepancy is extended to the full state using RBF interpolation. Relative to the no-interpolation case in (a), RBF interpolation substantially improves performance for sampling policies whose observations are either highly localized or do not directly cover the downstream convective evolution (most notably Top-RHS and Random), by spatially propagating sparse corrections to neighboring regions and effectively increasing the “influence radius” of each observation. In contrast, DEIM+KDE shows a more moderate change, suggesting that its KDE-enriched set already provides some spatial coverage around dynamically active regions, so additional smoothing yields a smaller marginal gain in this particular BFS setup. Uniform sampling remains the least effective even with RBF interpolation, reinforcing that, for this problem, where observations are placed can be at least as important as how the sparse corrections are spatially extended.

\section{Conclusion}
We presented a DEIM-based framework for interpretable diagnostics and targeted data assimilation of pre-trained Neural Ordinary Differential Equations (NODEs). By applying a time-windowed DEIM procedure to RHS snapshots, we extract trajectories of dynamically representative sampling locations and use them as an interpretable lens to analyze when and how NODE rollouts fail under extrapolative flow configurations. Building on this diagnostic signal, we introduced a DEIM-guided nudging strategy that allocates a limited assimilation budget to DEIM-identified locations enriched via kernel density estimation (KDE), enabling sparse yet dynamically informed corrections during NODE rollouts.

Through systematic experiments on two-dimensional vortex-merging and backward-facing step (BFS) flows, we demonstrated that DEIM trajectory analysis provides a physically interpretable diagnostic for extrapolation robustness. In vortex-merging extrapolation cases, the DEIM sampling trajectories extracted from the NODE RHS exhibit rapid structural collapse and irregular evolution, accompanied by accelerated growth of relative $L_2$ errors. These behaviors provide interpretable evidence that the trained NODE perceives such configurations as more extrapolative, revealing a direct connection between learned dynamics and underlying flow regimes. In the BFS flow, DEIM sampling points reveal a distinct failure mode associated with advection-dominated separated dynamics: while the ground-truth DEIM sampling points track vortices shed from the step and convected downstream, the trajectories extracted from the NODE prediction drift away from these coherent structures and collapse toward the inlet region as rollout errors accumulate, highlighting difficulties in preserving convective transport over long horizons.

Beyond interpretability, we showed that DEIM analysis can be leveraged to improve predictive performance through DEIM-guided nudging-based data assimilation. In vortex-merging experiments, nudging at DEIM+KDE locations substantially improves long-horizon stability and accuracy under sparse observations and helps restore the structural coherence of DEIM trajectories, indicating improved alignment between learned and true dynamics. In BFS, however, we found that the effectiveness of sparse nudging is strongly regime dependent: because windowed-DEIM reflects a lagged summary of recent dynamics, it can under-emphasize rapidly convecting, intermittently forming structures in advection-dominated separated flows, making instantaneous criteria (e.g., Top-RHS) or more spatially distributed policies (e.g., Random) competitive or even preferable under the same nudging budget. Moreover, due to the non-smooth hard-selection pipeline used to construct the nudging operator (greedy DEIM indices and density/top-$k$ enrichment), BFS rollouts exhibit pronounced late-time sensitivity to small changes in the selected point set; reporting ensemble mean and variance therefore provides a more faithful characterization of performance in this regime.

Overall, this work positions DEIM as a unified framework for interpretable diagnostics and targeted correction of neural differential equation models, while also highlighting that optimal sparse assimilation policies can be flow-regime dependent. Future directions include designing smoother (or differentiable) selection operators to reduce sampling-induced sensitivity, developing hybrid sampling strategies that combine structure-aware and instantaneous criteria for advection-dominated flows, and extending the framework to more complex turbulent configurations and realistic observation models.

\section{Acknowledgment}
This research used resources of the Argonne Leadership Computing Facility, which is a U.S. Department of Energy Office of Science User Facility operated under contract DE-AC02-06CH11357. HK and RM also acknowledge the support of computing resources through an AI For Science allocation at the National Energy Research Scientific Computing Center (NERSC) and from Penn State Institute for Computational and Data Sciences (ICDS). RM acknowledges the support of a YIP from ARO Modeling of Complex Systems (PM: Robert Martin).

\bibliography{bib}

\end{document}